\documentclass{article}
\usepackage{arxiv}

\usepackage[utf8]{inputenc} % allow utf-8 input
\usepackage[T1]{fontenc}    % use 8-bit T1 fonts

\usepackage{framed,multirow}

\usepackage{amssymb}
\usepackage{amsmath}
\usepackage{latexsym}
\usepackage{hyperref}
\usepackage{breakurl}
\usepackage{enumitem}
\usepackage{wasysym}
\usepackage{natbib}

\usepackage{xcolor}
\usepackage{xargs}
\usepackage[colorinlistoftodos,prependcaption,textsize=small]{todonotes}
\usepackage[framemethod=tikz]{mdframed}

\newcommandx{\unsure}[2][1=]{\todo[linecolor=red,backgroundcolor=red!25,bordercolor=red,caption={},#1]{#2}}
\newcommandx{\change}[2][1=]{\todo[linecolor=blue,backgroundcolor=blue!25,bordercolor=blue,caption={},#1]{#2}}
\newcommandx{\info}[2][1=]{\todo[linecolor=green,backgroundcolor=green!25,bordercolor=green,caption={},#1]{#2}}
\newcommandx{\improvement}[2][1=]{\todo[linecolor=purple,backgroundcolor=purple!25,bordercolor=purple,caption={},#1]{#2}}
\newcommandx{\discussion}[2][1=]{\todo[linecolor=blue,backgroundcolor=yellow!25,bordercolor=yellow,caption={},#1]{#2}}

\newcommandx{\thiswillnotshow}[2][1=]{\todo[disable,#1]{#2}}

\usepackage{url}
\usepackage{xcolor}
\usepackage{hyperref}
\usepackage{algorithm2e}
\usepackage{subcaption}

\definecolor{newcolor}{rgb}{.8,.349,.1}
\definecolor{peachfuzz}{RGB}{255, 190, 152}

\SetKwInput{KwData}{Input}
\SetKwInput{KwResult}{Output}

\newcommand{\themodel}{\ensuremath{\mathsf{CAP}}\xspace}
\newcommand{\themodelbf}{\ensuremath{\mathbf{CAP}}\xspace}

\title{CAP: Detecting Unauthorized Data Usage in Generative Models via Prompt Generation}

\author{
    Daniela Gallo \\
    ICAR - CNR and \\
    University of Salento \\
    Italy\\
    \texttt{daniela.gallo@icar.cnr.it} \\
    \And
    Angelica Liguori \\
    ICAR - CNR \\
    Italy \\
    \texttt{angelica.liguori@icar.cnr.it} \\
    \And
    Ettore Ritacco \\
    University of Udine \\
    Italy \\
    \texttt{ettore.ritacco@uniud.it}
    \And 
    Luca Caviglione \\
    IMATI - CNR \\
    Italy \\
    \texttt{luca.caviglione@cnr.it}\\
    \And
    Fabrizio Durante \\
    University of Salento\\
    Italy \\
    \texttt{fabrizio.durante@unisalento.it}
    \And
    Giuseppe Manco \\
    ICAR - CNR \\
    Italy \\
    \texttt{giuseppe.manco@icar.cnr.it} \\
}

\renewcommand{\shorttitle}{}

\begin{document}

\maketitle

\begin{abstract}
To achieve accurate and unbiased predictions, Machine Learning (ML) models rely on large, heterogeneous, and high-quality datasets. However, this could raise ethical and legal concerns regarding copyright and authorization aspects, especially when information is gathered from the Internet. With the rise of generative models, being able to track data has become of particular importance, especially since they may (un)intentionally replicate copyrighted contents. Therefore, this work proposes \textbf{Copyright Audit via Prompts generation} (\themodelbf\/), a framework for automatically testing whether an ML model has been trained with unauthorized data. Specifically, we devise an approach to generate suitable keys inducing the model to reveal copyrighted contents. To prove its effectiveness, we conducted an extensive evaluation campaign on measurements collected in four IoT scenarios. The obtained results showcase the effectiveness of \themodel\/, when used against both realistic and synthetic datasets. 

\end{abstract}

\keywords{Intellectual Property  \and Generative AI \and Model Attack \and Transformer}

%% main text
\section{Introduction}
\label{sec:introduction}

The success of modern Machine Learning (ML) systems depends on the quality and quantity of data used for training, which directly influences model performance and generalization capabilities. To this aim, high-quality, diverse, and representative datasets are essential for accurate and unbiased predictions. For instance, insufficient or biased data can lead to poor model performance, inaccuracies, and unintended consequences. Ethical and legal aspects are critical, too. In fact, data used to train ML models should be protected by copyright and proper authorizations for its usage should be granted on a case-by-case basis \citep{meurisch2021data}. 

With the widespread adoption of generative models, this issue has become more evident, especially since their ability to solve specific tasks often requires to train ML frameworks on large datasets embracing a wide range of sources. Unfortunately, data gathered from Internet may be subject to privacy policies or constraints. Since models are not able to discriminate among public or ``restricted'' sources, they may generate outputs that potentially replicate content without proper authorization~\citep{Li2024DiggerDC}. As an example, the ubiquitous diffusion of cost-effective IoT devices may lead to uncontrolled data ingestion campaigns for profiling users or industrial processes~\citep{jin2018they}. 
Thus, unauthorized usages and leakages of ML-generated contents should be promptly discovered and proven~\citep{sobel2017artificial}. An important research area considers how to test ML applications, e.g., to assess their correctness when used in mission-critical domains. Compared to classical software, ML pipelines utilize larger execution spaces requiring suitable tools to automatically test behaviors~\citep{braiek2020testing} or supporting libraries~\citep{gu2022muffin}.

In this perspective, our work addresses the problem of testing whether a black-box generative model has been trained with unauthorized data. Inquiring whether data has been used or not for training an ML model is known as \textit{membership inference} problem. However, different from classical Membership Inference Attacks (MIAs) \citep{shokri2017membership} that directly check if a given slice of information has been used in the training phase, we cannot directly inspect the training set used by the generative model, as only the owner knows it. To this aim, we propose \textbf{Copyright Audit via Prompts generation} (\themodelbf\/), a framework that can generate suitable keys (prompts) that induce the generative model to produce copyrighted values, allowing to detect unauthorized use of information. Since testing an ML model could be time-consuming, we also define a framework that is computationally optimized. Specifically, we propose a speedup procedure to enhance the performance of our approach in areas where it already demonstrates proficiency.

To prove the effectiveness of \themodel\/, we evaluate whether measurements of IoT sensors have been used without consent. Such a scenario is of paramount importance, since the pervasive adoption of IoT devices to control industrial deployments or drive urban intelligence frameworks may lead to severe privacy leaks or unwanted user profiling \citep{7902207}. We consider two main use cases. The first takes into account supervisory control and data acquisition settings, where measurements guarantee the functioning of a machinery and its maintenance, as well as the energy consumption of a vast urban area. The second bears with body sensors for tracking the head posture, e.g., for health or immersive applications.

The contributions of this work can be summarized as follows: (1) a framework for creating prompts to explore the training set; (2) a specific speedup strategy for its training; and (3) a performance evaluation on both realistic and synthetic datasets.

The rest of the paper is structured as follows. Section \ref{sec:related_works} reviews past research on membership inferences and similar attacks, while Section \ref{sec:contribution} presents our approach for generating prompts able to reveal the use of unauthorized data during the training phase. Section \ref{sec:experimental_assessment} showcases numerical results collected by considering realistic and synthetic datasets, and Section \ref{sec:conclusions} concludes the paper and hints at future research directions. 

\section{Related Work}
\label{sec:related_works}

Generating outputs able to highlight copyright infringement requires appropriate prompts or keys, which are often challenging to acquire, especially for real-world scenarios. Therefore, we aim to generate them, bridging the gap between theoretical constructs and practical applications, thus enhancing the utility and robustness of the model when deployed in realistic settings. However, the literature primarily focuses on the issue of copyright violation and data rights preservation~\citep{Ren2024CopyrightPI}.

In this perspective, a major corpus of works deals with how Deep Neural Networks (DNNs) tend to overfit the training data, retaining specific information. This can be leveraged by a threat actor to implement membership inference schemes~\citep{shokri2017membership}, allowing to distinguish between members and non-members data. To this aim, MIAs exploit the tendency of the model to exhibit higher confidence and lower loss on training samples compared to unseen ones. In binary classifier-based MIA~\citep{shokri2017membership}, attackers use shadow training to create shadow models mimicking the target one. Shadow models are trained on data similar to the training set of the target model, and their prediction vectors are labeled as ``member'' or ``non-member.'' Obtained labeled data is then used to train a binary classifier to recognize differences of the model when exposed to members and non-members. 
A more complex approach relies on indicators like prediction correctness, loss, confidence, and entropy, which are compared against suitable thresholds to infer membership~\citep{Yeom2017PrivacyRI, salem2019model}.

Different from our scenario, MIA schemes require substantial auxiliary information. Specifically, attackers need a non-negligible knowledge of the architecture and parameters of the target model (for \textit{white-box} attacks)~\citep{8835245} or the access to output predictions (for \textit{black-box} attacks)~\citep{10.1145/3372297.3417238}. They also need auxiliary datasets and computational resources for training shadow models or conducting metric-based evaluations.
Mitigation strategies against MIAs include regularization techniques, dropout, and differential privacy. Regularization techniques such as L2 regularization~\citep{shokri2017membership} can help reduce overfitting, making it harder for attackers to distinguish between members and non-members. Dropout~\citep{salem2019model}, which involves randomly dropping units during the training process, also helps to ``obfuscate'' the model behavior by preventing the model from relying too heavily on specific features of the training data. Lastly, differential privacy~\citep{LOGAN} introduces some noise into the training process, further protecting with a single-point granularity data from being identifiable.

Therefore, to the best of our knowledge, there are not any relevant attempts aiming at generating keys or prompts to induce the model to leak unauthorized data. Such a process resembles fuzzing testing~\citep{zhu2022fuzzing}, which has already been investigated for creating adversarial examples~\citep{guo2018dlfuzz} but never for prompts able to infer information of a dataset.

\section{Exposing Data through Prompt Generation}
\label{sec:contribution}
We introduce \themodel\/ by setting the notation and formally devising the problem statement. In the following, we denote the data to be identified by using the terms \textit{copyrighted} and \textit{unauthorized} in an interchangeable manner, unless doubts arise.

\begin{figure}
\centering
   \subfloat[Let $\Phi$ be a publicly-available generative model trained on a private dataset $\mathcal{D}_1$. This dataset contains instances that may be protected by copyright (represented as circles) and examples that are not protected by copyright (represented as squares). The model is designed to generate, starting with specific prompts (keys), realistic synthetic data based on this training set. Consequently, it might reproduce or closely resemble the copyrighted instances (circles), potentially violating copyright laws.\label{fig:toy_example_a}]{
   \includegraphics[width=.35\linewidth]{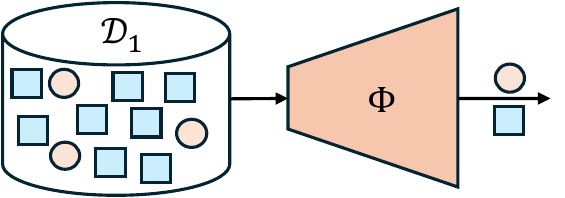}
   }
   \qquad \quad
   \subfloat[Let $\mathcal{D}_2$ be a dataset with copyrighted examples (represented as circles and stars). The data owner suspects that the generative model $\Phi$ has used its data without permission. To verify this, the owner uses another generative model, $\Theta$, to generate prompts (represented as triangles), triggering the generation through $\Phi$, which is only used in inference mode, to produce data similar to the dataset in question. If $\Phi$ generates data that matches any elements in $\mathcal{D}_2$ (circles), it indicates that the model is suspected of being trained on the copyrighted data. Conversely, if it produces only different data (pentangles), it suggests that $\Phi$ did not use the dataset of the owner and, therefore, did not infringe on the copyright.\label{fig:toy_example_b}]{
   \includegraphics[width=.55\linewidth]{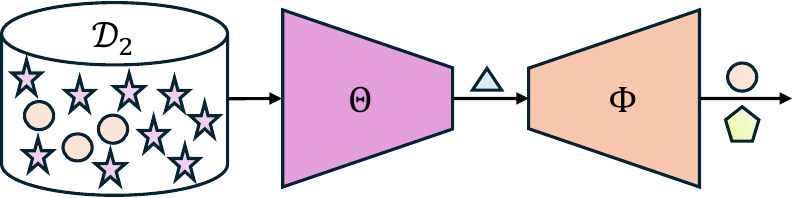}
   
   }
   \caption{An example of the reference scenario.}
   \label{fig:toy_example}
\end{figure}

Let $\mathcal{D}_1$ be a set of key-value pairs $(k, v)$, where $k \in K$ and $v \in V$. Suppose $\mathcal{D}_1$ is a confidential training set used to train~$\Phi$, a model designed to generate elements from $V$ in response to inputs from $K$. Additionally, assume the existence of a copyrighted dataset $\mathcal{D}_2 \subseteq V$, whose values require explicit authorization for use. If the owner of $\mathcal{D}_2$ suspects that their data was used without permission in the training of $\Phi$, they need to determine whether $\mathcal{D}_1$ contains any elements from $\mathcal{D}_2$, i.e., whether $\{v \mid (k, v) \in \mathcal{D}_1\} \cap \mathcal{D}_2 \neq \emptyset$.
The challenge lies in verifying whether $\Phi$ was trained on $\mathcal{D}_2$ despite the impossibility of directly inspecting the secret training set $\mathcal{D}_1$.
An example of the reference scenario is depicted in Figure~\ref{fig:toy_example_a}.

To address this problem, we propose the framework illustrated in Figure~\ref{fig:toy_example_b}. The idea is to exploit another generative model, $\Theta$, to produce specific prompts that induce $\Phi$ to generate elements identical or significantly similar to a subset of $\mathcal{D}_2$.
Specifically, given a value $v \in \mathcal{D}_2$, we aim at inferring a key $k\in K$, using $\Theta$, which attempts to force $\Phi$ to generate either an exact copy or a slightly altered variant of $v$. Formally, the framework is defined by Algorithm~\ref{alg:violations}, which allows finding violations, e.g., whether the data has been used without consent.

\RestyleAlgo{ruled}
\begin{algorithm}
\caption{Finding violations}\label{alg:violations}
\LinesNumbered
\KwData{\quad Copyrighted dataset  $\mathcal{D}_2$, \\
\qquad \qquad  Generative Model $\Phi$ trained on $\mathcal{D}_1$, \\
\qquad \qquad Prompt Generator $\Theta$ trained on $\mathcal{D}_2$, \\
\qquad \qquad Distance function $\Delta$, \\
\qquad \qquad Tolerance $\delta$.}
\KwResult{\, Set of violations $\mathcal{V}$, \\
\qquad \qquad Set of prompts $\mathcal{K}$ triggering violations. }
$\mathcal{V} \gets []$\;
$\mathcal{K} \gets []$\;
\For{$v \in \mathcal{D}_2$} {
    $k \sim p_\Theta(\cdot | v)$\;
    $\hat{v} \sim p_\Phi(\cdot | k)$\;
    \If{$\Delta(\hat{v}, v) < \delta$}{
        $\mathcal{V}.\mathtt{append}(v)$\;
        $\mathcal{K}.\mathtt{append}(k)$\;
    }
}
\end{algorithm}

The algorithm is provided with $\mathcal{D}_2$, the pre-trained models $\Phi$ and $\Theta$, a suitable distance function $\Delta$, and a tolerance threshold~$\delta$.
The output of the algorithm is a pair of sets $(\mathcal{V}, \mathcal{K})$, containing, respectively, the values in $\mathcal{D}_2$ that $\Phi$ is able to replicate, thereby violating copyright, and the prompts generated by $\Theta$ that induce the generation of those replicas.
For each element $v \in \mathcal{D}_2$, we use $\Theta$ to sample a prompt $k$ that triggers $\Phi$ to generate $\hat{v}$. If the distance $\Delta$ between $\hat{v}$ and $v$ is less than the threshold~$\delta$, we assume that $\hat{v}$ and $v$ represent the same entity, providing evidence of copyright violation. We then store $v$ and $k$ in $\mathcal{V}$ and $\mathcal{K}$, respectively.
The choice of $\Delta$ and $\delta$ depends on the specific scenario of interest and the interpretation of copyright violation. For instance, they might compare the entire information in $\hat{v}$ and $v$, focus on the closeness of certain portions of them, or search for the similarity of specific features.

\subsection{Training the Prompt Generator}
\begin{figure}[t]
    \centering
    \includegraphics[width=0.65\linewidth]{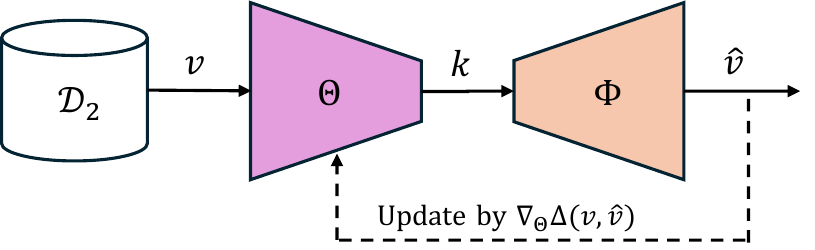}
    \caption{Training Process of the Prompt Generator. The model $\Theta$ generates keys $k$ when provided with values $v$ that we want to inspect. The pre-trained model~$\Phi$ then produces values $\hat{v}$ in response to the inputs $k$. Finally, the model~$\Theta$ is updated to minimize the distance $\Delta$ between $v$ and $\hat{v}$.}
    \label{fig:training}
\end{figure}

To find unauthorized usages of data or copyrighted information, we need to train our Prompt Generator $\Theta$. Its training process is illustrated in Figure~\ref{fig:training} and formalized in Algorithm~\ref{alg:training}.

\RestyleAlgo{ruled}
\begin{algorithm}[t]
\caption{Training $\Theta$}\label{alg:training}
\LinesNumbered
\KwData{\quad Copyrighted dataset $\mathcal{D}_2$, \\
\qquad \qquad  Generative Model $\Phi$ trained on $\mathcal{D}_1$, \\
\qquad \qquad Distance function $\Delta$.}
\KwResult{\, Prompt Generator $\Theta$}
$\Theta \gets \mbox{random initialization}$\;
\While{$\Theta$ has not converged}{
    Sample mini-batch of $m$ values $\{v_1, \ldots, v_m\}$ from $\mathcal{D}_2$\;
    Update $\Theta$ by descending the stochastic gradient: \\
        \qquad $\nabla_\Theta \frac{1}{m} \sum_{i=1}^m \Delta\left(v_i, \Phi\left(\Theta\left(v_i\right)\right)\right)$
}
\end{algorithm}

The algorithm takes as input the copyrighted dataset $\mathcal{D}_2$, the model $\Phi$ pre-trained on the dataset $\mathcal{D}_1$, and the distance function $\Delta$. The output of the algorithm is the trained model $\Theta$, whose weights are adjusted to generate prompts that try to induce $\Phi$ to produce values closely matching those in $\mathcal{D}_2$. Specifically, starting with random initialization, the model $\Theta$ is iteratively updated through a mini-batch-based optimization process until convergence.
Given $\Phi$ and $v \in \mathcal{D}_2$, we propose a loss function for $\Theta$ to minimize that penalizes based on the distance between the input and the output:
\begin{equation}
    \mathit{loss}_\Theta(v) = \Delta\left(v, \Phi\left(\Theta\left(v\right)\right)\right) ,
\end{equation}
where $\Theta(v)$ produces the prompt $k$ that triggers $\Phi$ to generate $\hat{v}$.
We remark that, in our scenario, the model $\Phi$ is pre-trained and used in inference mode. This makes the framework suitable for typical black-box membership inference approaches, where the model is not fully disclosed and the problem is to gather evidence of unauthorized exploitation of proprietary data.

\subsection{Speeding up the training of the Prompt Generator}

The models we consider are basically encoder-decoder architectures. These include, e.g., Diffusion Models \citep{DBLP:conf/icml/Sohl-DicksteinW15,DBLP:conf/nips/HoJA20}, Generative Adversarial Networks \citep{DBLP:journals/cacm/GoodfellowPMXWO20}, or Variational Autoencoders \citep{Kingma2014}. Furthermore, they typically employ components based on transformers~\citep{VaswaniSPUJGKP17}. As a result, training $\Theta$ within the proposed framework can be prohibitively expensive. 
To address this issue, we propose a speedup procedure based on the idea of enhancing the performance of the model in areas where it already demonstrates proficiency. Contrary to the traditional principle of ``learning from mistakes'', here $\Theta$ is designed to focus on identifying a subset of data that indicates unauthorized usages. 
The goal is not to generalize across all data but to concentrate on the data where the model is most confident such as those that minimize the error.
The proposed method removes from the training set those data points that the model identifies as noise, i.e., those with higher generalization errors. By doing so, the model can streamline its training process by focusing on more relevant and informative data. This selective approach reduces the overall training time while maintaining or enhancing the performance of the model over the filtered subset of the training data.
This procedure is applied when the generator no longer learns new information from the data, resulting in a stagnating or not improving loss.

To this aim, we adopt a robust approach by leveraging the Generalized Pareto Distribution (GPD) \citep{vignotto2020extreme} to estimate the tail of the error distribution. According to the Pickands–Balkema–De Haan theorem \citep{b3094dd8-e768-3594-9f6b-4c6434a91f76}, the conditional distribution of a random variable $X$ given a high threshold $u$, i.e., $(X\mid X>u)$, converges to the GPD as $u\to +\infty$. This is true when the law of $X$ belongs to a variety of families, like exponential distributions (e.g., Gaussian and Laplacian), stable distributions (e.g., Cauchy and Levy), and power law distributions (Student-t and Pareto). 

We exploit this property within \themodel\/ by devising a data reduction process that involves two steps. First, we fit a generalized Pareto distribution to the empirical error data. This involves estimating the parameters of the GPD that best represent the tail behavior of the error distribution. Second, we determine an appropriate threshold level based on the fitted GPD. In this context, we adopt the $80$th percentile as the reference threshold. This choice is guided by a heuristic application of the Pareto principle \citep{wilkinson2006revising}, which posits that approximately $20\%$ of the causes are responsible for $80\%$ of the effects.
The entire process is formalized in Algorithm~\ref{alg:training_speed_up}.

\RestyleAlgo{ruled}
\begin{algorithm}[ht!]
\caption{Optimized training $\Theta$}\label{alg:training_speed_up}
\LinesNumbered
\KwData{\quad Copyrighted dataset $\mathcal{D}_2$, \\
\qquad \qquad  Generative Model $\Phi$ trained on $\mathcal{D}_1$, \\
\qquad \qquad Distance function $\Delta$, \\
\qquad \qquad Patience value $\alpha$, \\
\qquad \qquad Tolerance for patience $\omega$.}
\KwResult{\, Prompt Generator $\Theta$}
$\Theta \gets \mbox{random initialization}$\;
$I \gets \{1, \ldots, |\mathcal{D}_2|\}$\;
\While{$\Theta$ has not converged}{
    $\mathit{errors} \gets []$\;
    $\mathit{indexes} \gets []$\;
    Split $\mathcal{D}_2[I]$ in mini-batches $\{\mathcal{B}_1, \mathcal{B}_2, \ldots\}$\;
    \For{$\mathcal{B} \in \{\mathcal{B}_1, \mathcal{B}_2, \ldots\}$}{
        $g \gets 0$\;
        \For{$v \in \mathcal{B}$}{
            $\varepsilon = \Delta\left(v, \Phi\left(\Theta\left( v \right)\right)\right)$\;
            $g \gets g + \varepsilon$\;
            $\mathit{errors}.\mathtt{append}(\varepsilon)$\;
            $\mathit{indexes}.\mathtt{append}(\mbox{\textit{index of} } v \mbox{\textit{ in }} I)$\;
        }
        Update $\Theta$ by descending the stochastic gradient: $\nabla_\Theta \frac{1}{|\mathcal{B}|} g$
    }
    \If{$|I| > |\mathcal{D}_2| / 3$ {\bf and} $\mathit{Patience}\left( \mathtt{mean}(\mathit{errors}), \alpha, \omega\right)$}{
        $\tau \gets \arg\min_{\varepsilon \in \mathit{errors}} |\mathtt{GPD}(\mathit{errors})_{80\%} - \varepsilon|$\;
        Sort both $\mathit{indexes}$ and $\mathit{errors}$ according to $\mathit{errors}$ descendently\;
        \For{$i \in \{1, \ldots, |\mathit{indexes}|\}$ {\bf and} $|I| > |\mathcal{D}_2| / 3$}{
            \lIf{$\mathit{errors}_i >= \tau$}{
                $I \gets I - \textit{indexes}_i$
            } 
        }
    }
}
\end{algorithm}

For speeding up the training, we consider the same input and output of the previous training algorithm (see Algorithm \ref{alg:training}). Moreover, the algorithm takes as input two additional parameters, which are the patience value~$\alpha$ and the tolerance for patience~$\omega$.
Specifically, we initialize a list of indexes of elements in $\mathcal{D}_2$. 
Then, the dataset $\mathcal{D}_2$ is split into mini-batches $\{\mathcal{B}_1, \mathcal{B}_2, \ldots\}$. For each mini-batch $\mathcal{B}$, we cumulatively collect within $\mathit{errors}$ the errors $\varepsilon$ resulting from the $\Delta$ loss function between the predicted value $\Phi(\Theta(v))$ and the actual value $v$, as well as descend the gradient on the batch loss $g$.  
Within the training procedure, a patience function is integrated, to manage when to drop data points based on the progression of the loss function. The idea of the patience mechanism is to avoid unnecessary iterations when there are no significant improvements in the loss, thus saving time and computational resources. For this, we keep track of the observed best loss value. If the current loss (namely, $\mathtt{mean}(\mathit{errors})$) does not improve above a minimum threshold $\omega$ for a specific number of iterations $\alpha$, and the size of the index set $I$ is above a predefined amount (one-third of the original dataset size in our framework), the algorithm cuts all the elements in $\mathcal{D}_2$ that are above threshold value $\tau$ that is closest to the $80$th percentile of the GPD fitted on the errors.

\begin{table*}[t]
    \centering
     \caption{Dataset Description}
    \resizebox{1.\textwidth}{!}{
   % \small
    \begin{tabular}{|l|c c c c | c c c c|}
        \hline
         \textbf{Datasets} & \# records & \# features & Sequence length & \# sequences & $\mathcal{D}_{tr}$ & $\mathcal{D}_{v}$ & $\mathcal{D}_{nc}$ & $\mathcal{D}_{c}$\\
         
         \hline

         \textbf{Pump Sensor} & 117,912 & 51 & 60 & 1,965 & 669 & 623 & 673 & 201\\
         \hline
         
         \textbf{Elevator Failure} & 93,882 & 8 & 60 & 1,564 & 486 & 483 & 595 & 146\\
         
        \hline
        \textbf{Electric Power Consumption} & 52,416 & 8 & 30 & 1,747 & 598 & 547 & 602 & 180\\
         
        \hline
        \textbf{Head Posture} & 44,992 & 9 & 30 & 1,499 & 452 & 452 & 595 & 136\\
        \hline

        \textbf{Synthetic} & 360,000 & 16 & 60 & 6,000 & 2,000 & 2,000 & 2,000 & 600 \\
        \hline

        \textbf{Synthetic-Overlap} & 360,000 & 16 & 60 & 6,000 & 2,000 & 2,000 & 2,000 & 600\\
        \hline
    \end{tabular}
    }
    \label{tab:dataset_descriptions}
\end{table*}

By setting the threshold at the $80$th percentile, we conservatively exclude the top $20$\% of errors, which are presumed to be the most impactful. This exclusion helps mitigating the adverse effects of these extreme errors on our analyses, ensuring more reliable and robust results. As a result, the approach balances the need to retain sufficient data for meaningful analysis while eliminating the most ``problematic'' data points.

Through a worst-case and best-case scenario analysis, we observe that in the worst-case scenario of the optimization process, the patience function enables to not shrink the training set, i.e., no elements are actually removed from the dataset. Since errors are small and the whole training set is useful to the model, no examples can be discarded. Owing to small errors, the model will still converge in a short time. In the best-case scenario, the optimization process discards data that are not useful for learning, allowing the model to proceed with training on a reduced dataset and taming training times. By considering both scenarios, we can claim that this approach ensures a balanced and effective data management strategy that positively affects the training process.

\section{Experimental Results}
\label{sec:experimental_assessment}

This section demonstrates the effectiveness of \themodel\/ when used against four different public datasets, capturing industrial- or carrier-grade scenarios where data theft is considered a relevant hazard. We aim to answer the following research questions:

\begin{itemize}[noitemsep]

    \item \textbf{RQ1}. Can \themodel\/ generate keys that actually detect copyright infringement?
    \item \textbf{RQ2}. How effective is the speedup procedure in detecting copyright infringement while also reducing training times?
    \item \textbf{RQ3}. What are the strengths and limitations of \themodel\/?
\end{itemize}

We first provide an overview of the datasets, the implementation details, and the evaluation protocol. Then, we will discuss the results and limitations.

\subsection{Datasets and Evaluation Protocol}
In principle, \themodel\/ can be used within any encoding-decoding scenario, including text and/or image generation from prompts. To test its effectiveness within simplified architectures that do not include large models (and hence prevent the need for excessive computational power), we restrict our attention to encoding-decoding of time series. The analysis of more complex (large) architectures is demanded to future work. The experiments focus on real and synthetic datasets described below.
\begin{itemize}[noitemsep]

        \item \textbf{Pump-Sensor}\footnote{Pump sensor data for predictive maintenance,  online: \url{www.kaggle.com/datasets/nphantawee/pump-sensor-data}} is a sequential dataset related to failures in a water pump in which the period of observation is of $5$ months, sampled each $1$ minute. The dataset consists of raw (numerical) values collected from $52$ sensors.

        \item \textbf{Elevator Failure}\footnote{Elevator predictive maintenance dataset, online: \url{www.kaggle.com/datasets/shivamb/elevator-predictive-maintenance-dataset}} is a set of observations obtained from several IoT sensors for predictive maintenance in the elevator industry. Each observation is sampled at $4$Hz in high-peak and evening elevator usage in a building. 

        \item \textbf{Electric Power Consumption}\footnote{Electric power consumption dataset, online: \url{www.kaggle.com/datasets/fedesoriano/electric-power-consumption}} is a collection of observations of energy consumption in the city of Tetouan, located in the north of Morocco. The data has been recorded every $10$ minutes. 
        
        \item \textbf{Head Posture}
        is a selection of time series from three inertial sensors, along with labels for various head postural motions. The relevant time series describing these motions are in the form of Roll, Pitch, and Yaw Euler angles~\citep{9280106,9315418}.

\end{itemize}
Since some datasets contained missing values or duplicate records, they undergo pre-processing to eliminate null values and/or duplicates.

To demonstrate both the applicability and the limitations of our approach, we also analyzed the behavior of \themodel\/ in a controlled environment via synthetically generated data.

In our experiments, based on the reference scenario from Section \ref{sec:contribution}, we create three balanced subsets from each real dataset using hierarchical clustering: $\mathcal{D}_{tr}$, $\mathcal{D}_{v}$, and $\mathcal{D}_{nc}$. The first two subsets are used to train and validate model $\Phi$, forming dataset $\mathcal{D}_1$. A small portion of $\mathcal{D}_{tr}$, called $\mathcal{D}_c$, is sampled as copyrighted data used by $\Phi$ without authorization. This data is combined with $\mathcal{D}_{nc}$ (copyrighted data unseen by $\Phi$) to form~$\mathcal{D}_2$.

Concerning synthesized data, we create group sequences adhering to distinct patterns. Each sequence is split into two parts: the first is generated from a multivariate Gaussian distribution with a specific mean, while the second is drawn from a different Gaussian distribution. 
As a result, $\mathcal{D}_{tr}$, $\mathcal{D}_{v}$, and $\mathcal{D}_{nc}$ contain the same number of samples obtained from sequences belonging to differentiated patterns (which are devised by different Gaussian parameters). 
Through this scheme, we generated two distinct datasets with a $1$-minute granularity. The first dataset (named \textbf{Synthetic} in the following), contains three subsets that are well-separated and have no overlapping elements, i.e., $\mathcal{D}_{v} \cap \mathcal{D}_{c} = \emptyset $ and $\mathcal{D}_{nc} \cap \mathcal{D}_{c} = \emptyset$.
The second dataset (\textbf{Synthetic-Overlap}) is characterized by overlapping Gaussian distributions that are used to generate both the first and third subsets. 
Table \ref{tab:dataset_descriptions} summarizes the statistics of the datasets.

\subsection{Implementation Details and Evaluation Metrics}
To implement \themodel\/, we used the PyTorch framework\footnote{
To ensure reproducibility, we have publicly released all the data and code necessary to replicate our experiments, see online at: \url{www.github.com/Angielica/WHAM-/tree/main/CAP}}. Both $\Phi$ and $\Theta$ are transformer-based models \citep{VaswaniSPUJGKP17}. We implemented the same architecture presented in the original paper \citep{VaswaniSPUJGKP17} and used the default parameters, specifically: the encoder and decoder are composed both of a stack of $6$ identical layers, while for the multi-head attention module, $8$ parallel attention heads are employed. The embedding layers produce outputs of dimension equal to $512$. 
We use the Adam optimizer 
with a learning rate equal to $1$e$-4$. Models $\Phi$ and $\Theta$ were trained for a maximum of $1,000$ and $500$ epochs, respectively. An early stopping procedure, with a patience of $30$, was applied. Additionally, during training, the model versions with the lowest validation loss (for $\Phi$) and the lowest training loss (for $\Theta$) were saved.

\begin{figure}
    \centering
    \includegraphics[width=0.65\columnwidth]{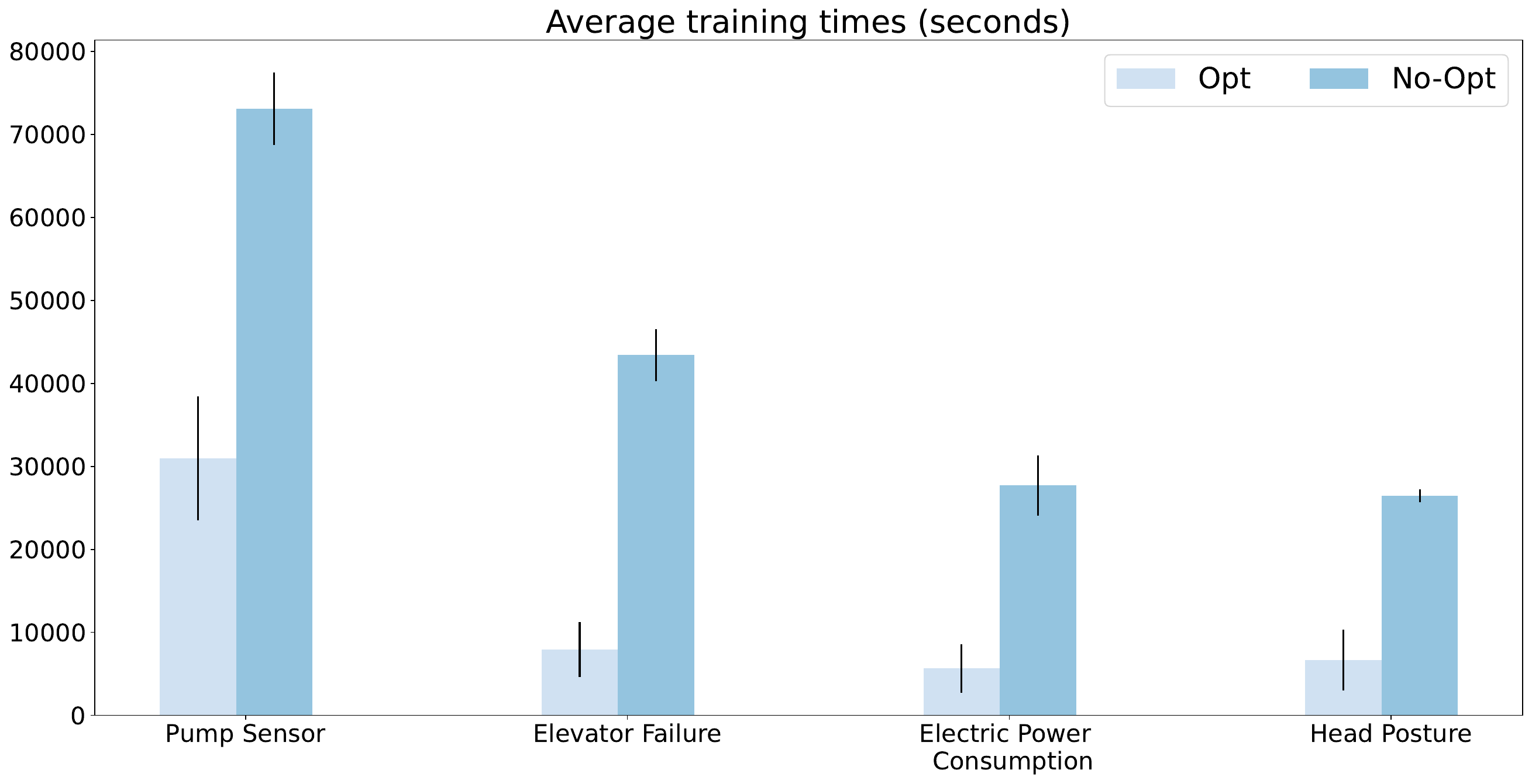}
    \caption{Running times with (Opt) and without (No-Opt) the speedup procedure.}
    \label{fig:running_times}
\end{figure}

To evaluate the performance of \themodel\/, we calculate the \textsl{Precision@K} metric that computes the number of times the copyrighted/unauthorized data is among the top $K$ data detected by the framework. The data are ranked based on the distance between $v \in \mathcal{D}_2$ and its generated version $\Phi(\Theta(v))$, sorted in descending order.
Additionally, we compute the Area Under the Cumulative Gains Curve (AUC-Gain), which quantifies the performance of the model in terms of having copyrighted instances early in the ranked list. All experiments are performed on $10$ runs, and the average values are reported, with statistical significance computed at $95\%$ confidence.

\subsection{Discussion of Results}

\begin{table*}
    \centering
    % \footnotesize
    \caption{Comparative analysis of the framework with (Opt) and without (No-Opt) the speedup procedure.}
    \resizebox{1.\linewidth}{!}{
    \begin{tabular}{|l|c c |c c| c c| c c | c c |}
        \hline
         \multirow{2}{*}{\textbf{Datasets}} & \multicolumn{2}{c|}{\textsl{Precision@5}}  & \multicolumn{2}{c|}{\textsl{Precision@10}} & \multicolumn{2}{c|}{\textsl{Precision@50}} & \multicolumn{2}{c|}{\textsl{Precision@100}} & \multicolumn{2}{c|}{AUC-Gain} 
         \\ 
         \cline{2-11}
         & No-Opt & Opt & No-Opt & Opt & No-Opt & Opt & No-Opt & Opt  & No-Opt & Opt
         \\
        
         \hline

         \textbf{Pump Sensor} & 100 $\pm$ 0  & 100 $\pm$ 0 & 100 $\pm$ 0 & 100 $\pm$ 0 & 98 $\pm$ 2  & 97 $\pm$ 3 & 91 $\pm$ 2 & 87 $\pm$ 3 & 0.95 $\pm$ 0.01 & 0.93 $\pm$ 0.01 
         \\
         \hline
         \textbf{Elevator Failure} & 100 $\pm$ 0  & 100 $\pm$ 0 & 100 $\pm$ 0 & 100 $\pm$ 0 & 52 $\pm$ 7  & 50 $\pm$ 7 & 55 $\pm$ 2 & 55 $\pm$ 2 &  0.92 $\pm$ 0.01 & 0.92 $\pm$ 0.01  
         \\
         \hline
         \textbf{Electric Power Consumption} & 74 $\pm$ 30  & 66 $\pm$ 28 & 68 $\pm$ 25 & 61 $\pm$ 23 & 54 $\pm$ 15  & 50 $\pm$ 15 & 48 $\pm$ 9 & 49 $\pm$ 8 & 0.74 $\pm$ 0.02 & 0.72 $\pm$ 0.02 
         \\
         \hline
         \textbf{Head Posture} & 100 $\pm$ 0  & 100 $\pm$ 0 & 100 $\pm$ 0 & 100 $\pm$ 0 & 100 $\pm$ 0  & 100 $\pm$ 0 & 100 $\pm$ 0 & 100 $\pm$ 0 & 0.96 $\pm$ 0.01 & 0.96 $\pm$ 0.01 
         \\
         \hline
    \end{tabular}}
    \label{tab:results}
\end{table*}

\begin{table*}
    \centering
    % \footnotesize
    \caption{Comparative analysis of the framework with Synthetic and Synthetic-Overlap data.}
    %\resizebox{1.\linewidth}{!}{
    \begin{tabular}{|l|c |c | c | c  | c | }
        \hline
         \multirow{1}{*}{\textbf{Datasets}} & \multicolumn{1}{c|}{\textsl{Precision@5}}  & \multicolumn{1}{c|}{\textsl{Precision@10}} & \multicolumn{1}{c|}{\textsl{Precision@50}} & \multicolumn{1}{c|}{\textsl{Precision@100}} & \multicolumn{1}{c|}{AUC-Gain} %& \multicolumn{2}{c|}{Running Times (sec)} 
         \\ 
        
         \hline
         \textbf{Synthetic} & 100 $\pm$ 0 & 100 $\pm$ 0 & 100 $\pm$ 0  & 100 $\pm$ 0 & 1 $\pm$ 0  % & 29,243 $\pm$ 12,363 & 13,260 $\pm$ 3,624
         \\
         \hline
         \textbf{Synthetic-Overlap} & 38 $\pm$ 29 & 36 $\pm$ 27 & 33 $\pm$ 27 & 31 $\pm$ 27 & 0.49 $\pm$ 0.19 \\
         \hline
    \end{tabular}%}
    \label{tab:syn_results}
\end{table*}

Table \ref{tab:results} reports the results of the evaluation with (Opt) and without (No-Opt) the speedup procedure. In response to RQ1, the findings indicate that \themodel\/ effectively generates keys that compel the model $\Phi$ to produce values used during its training, thereby uncovering copyright infringement. Notably, in almost all the real-world datasets, the copyrighted data consistently rank within the top $5$ and $10$ positions showing a precision equal or near to $100$$\%$. By increasing the value of $K$, we can observe a degradation of the performance.
Furthermore, the model achieves an AUC-Gain close to $1$, demonstrating its high effectiveness in distinguishing between copyrighted and non-copyrighted instances. In other words, the model is nearly optimal in quickly identifying copyrighted data. 
To answer RQ2, we compare the performance and the running times of \themodel\/~with and without the speedup strategy. Specifically, the time optimization procedure proves to be highly effective in lowering training times without compromising the ability to detect copyright infringement. As reported in Table \ref{tab:results}, all metrics show minimal to no degradation when the optimization is applied. On the contrary, as illustrated in Figure \ref{fig:running_times}, the running times are substantially reduced across all datasets when the procedure is implemented. 

\begin{figure}
\centering
   \subfloat[t-SNE for Head Posture dataset.\label{fig:head_tsne}]{
   \includegraphics[width=.45\linewidth]{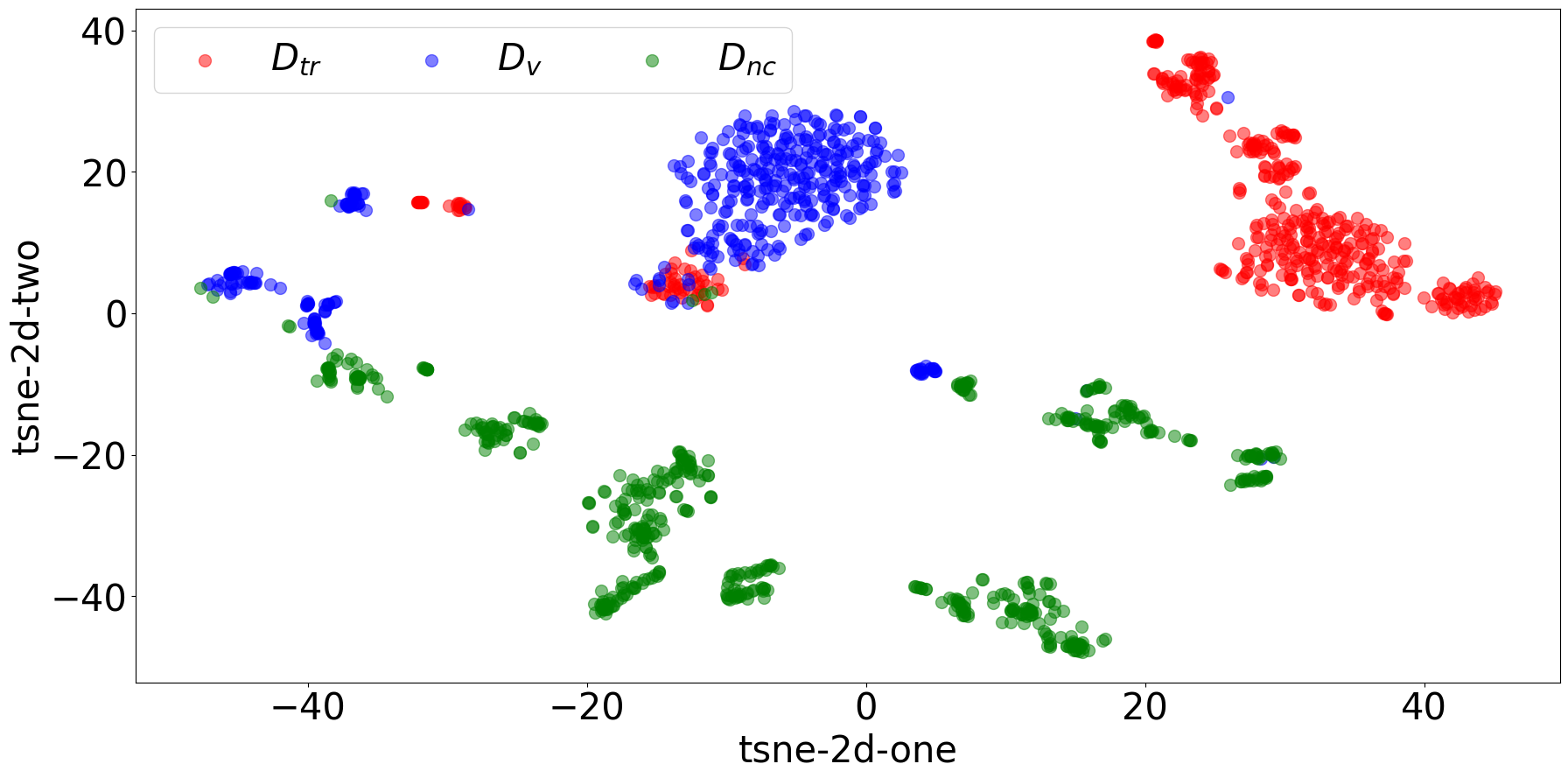}
   }
   \qquad \quad
   \subfloat[t-SNE for Power Consumption dataset.\label{fig:power_tsne}]{
   \includegraphics[width=.45\linewidth]{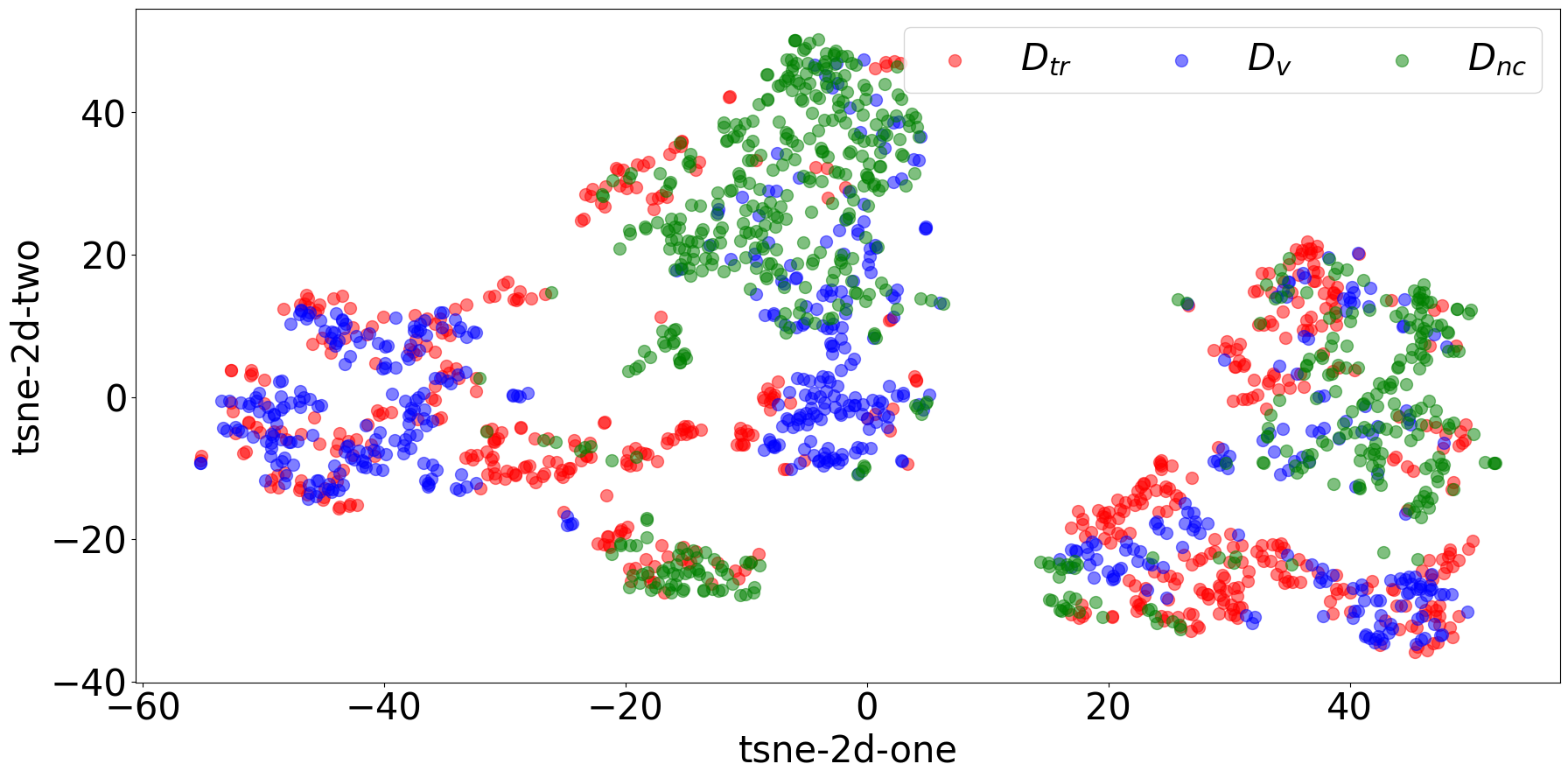}
   
   }
   \caption{t-SNE for datasets with different characteristics.}
   \label{fig:tsne}
\end{figure}

We point out that, \themodel requires that the information to be checked exhibits traits of uniqueness, i.e., its distribution is specific enough to be considered representative of copyrighted material. Without this distinctiveness in the data, the model may struggle to accurately differentiate between ``original'' and ``derivative'' content. In such cases, it could either incorrectly attribute copyright to a generic piece of information or fail to distinguish between distributions that vary only slightly.
This is illustrated in Figures~\ref{fig:head_tsne} and~\ref{fig:power_tsne}, which depict a two-dimensional representation of datasets with different characteristics. The first dataset in Figure~\ref{fig:head_tsne} is characterized by non-overlapping distributions, whereas Figure~\ref{fig:power_tsne} showcases data characterized by two overlapping (and hence indistinguishable) slices of data. The behavior of \themodel in these situations is illustrated in Table~\ref{tab:results}. In fact, the low performance of the model on the \textit{Electric Power Consumption} can be explained by the overlap between the copyrighted and non-copyrighted sets.

To further strengthen this aspect and answer RQ3, we have also conducted in-vitro experiments. The results, reported in Table~\ref{tab:syn_results}, show that on the \textit{Synthetic-Overlap} dataset, where the distributions of the copyrighted and non-copyrighted sets overlap, the model struggles to accurately identify copyrighted data due to their similarity to the non-copyrighted ones. In contrast, on the \textit{Synthetic} dataset, where data is generated using separate distributions, for all the values of $K$, the model achieves perfect scores. Additionally, an AUC-Gain of $1.00$ demonstrates the ability of \themodel to distinguish between copyrighted and non-copyrighted data perfectly.

% \begin{figure}[t]
%     \centering
%     \includegraphics[width=0.8\columnwidth]{head_posture_tsne.png}
%     \caption{t-SNE for Head Posture dataset.}
%     \label{fig:head_tsne}
% \end{figure}

% \begin{figure}[t]
%     \centering
%     \includegraphics[width=0.8\columnwidth]{power_consumption_tsne.png}
%     \caption{t-SNE for Power Consumption dataset.}
%     \label{fig:power_tsne}
% \end{figure}

\section{Conclusions}
\label{sec:conclusions}

In this paper, we presented \themodel, a framework for generating suitable prompts to discover whether a given information belongs to the training set. Since exploring large datasets to reveal unauthorized usages could be time consuming, we introduced an optimized generation procedure. To prove the effectiveness of \themodel\/, we conducted tests against realistic and synthetic datasets, which allow to consider an ML model taking advantage of leaked/stolen data. Results demonstrated the effectiveness of our approach. Specifically, in scenarios where there is a clear separation between copyrighted and non-copyrighted data, the model can perfectly identify all the copyrighted data. 

Future work aims at extending the framework to discover more standard copyright violations, e.g., in text or visual contents generated via complex and large ML models, such as diffusion models and large language models.

\section*{Acknowledgments}
This research was partially funded by Project WHAM! - Watermarking Hazards and novel perspectives in Adversarial Machine learning (B53D23013340006), by Project RAISE - Robotics and AI for Socio-economic Empowerment (ECS00000035), by Project STRIVE/URAN - Advanced Approaches for Transitions in Urban Environments, and by the Project SERICS (PE00000014) under the NRRP MUR program funded by the European Union – Next Generation EU.

\bibliographystyle{plainnat}
%\bibliography{refs}

\end{document}